\documentclass[journal]{IEEEtai}
\usepackage{lineno,hyperref}
\usepackage{amsmath}
\usepackage{color}
\usepackage{amssymb}
\modulolinenumbers[5]
\usepackage{graphicx}
\usepackage{caption}
\usepackage{multirow}
\usepackage{bbding}

\ifCLASSINFOpdf
\else
\fi
\begin{document}
%
\title{Global Attention-Guided Dual-Domain Point Cloud Feature Learning for Classification and Segmentation}
%
%
%

\author{Zihao Li,
        Pan Gao,
        Kang You,
        Chuan Yan,
        and Manoranjan Paul
\thanks{This work was supported by  the Natural Science Foundation of China under Grant 62272227. \emph{(Corresponding author: Pan Gao.)}}
\thanks{Z. Li, P. Gao and K. You are with College of Artificial Intelligence, Nanjing University of Aeronautics and Astronautics, Nanjing 211106, China. (pride\_19@163.com, gaopan.1005@gmail.com, youkang@nuaa.edu.cn)}
\thanks{C. Yan is with Creativity and Graphics Lab, George Mason University, Virginia 22030, US. (cyan3@gmu.edu)}
\thanks{M. Paul is with School of Computing, Mathematics, and Engineering, Charles Sturt University, NSW 2678, Australia. (mpaul@csu.edu.au)}
}

%
%

\markboth{Journal of \LaTeX\ Class Files,~Vol.~14, No.~8, August~2015}%
{Shell \MakeLowercase{\textit{et al.}}: Bare Demo of IEEEtran.cls for IEEE Journals}
%




\maketitle

{
\begin{abstract}
    Previous studies have demonstrated the effectiveness of point-based neural models on the point cloud analysis task. However, there remains a crucial issue on producing the efficient input embedding for raw point coordinates. Moreover, another issue lies in the limited efficiency of neighboring aggregations, which is a critical component in the network stem. In this paper, we propose a Global Attention-guided Dual-domain Feature Learning network (GAD) to address the above-mentioned issues. We first devise the Contextual Position-enhanced Transformer (CPT) module, which is armed with an improved global attention mechanism, to produce a global-aware input embedding that serves as the guidance to subsequent aggregations. Then, the Dual-domain K-nearest neighbor Feature Fusion (DKFF) is cascaded to conduct effective feature aggregation through novel dual-domain feature learning which appreciates both local geometric relations and long-distance semantic connections. Extensive experiments on multiple point cloud analysis tasks (e.g., classification, part segmentation, and scene semantic segmentation) demonstrate the superior performance of the proposed method and the efficacy of the devised modules.
\end{abstract}
}

\begin{IEEEImpStatement}

The 3D point cloud is a fundamental data structure utilized in a wide range of applications, such as intelligent manufacturing, automatic driving, robot control, etc. The efficacy of these applications fundamentally depends on their ability to accurately capture and understand the geometric information of the point cloud data. While most existing methods tend to use complex and expensive frameworks to enhance the understanding ability of the neural model, we propose a simple and cost-efficient method for better information extraction. {Experiments show that the proposed method demonstrates state-of-the-art performance with the adaptable traits present in the proposed modules, which has the potential to aid in other point cloud analysis tasks (e.g., completion and compression). In addition, our fully armed method demonstrates a 4\% improvement in classification compared to the baseline in the ablation study, marking a significant leap forward.}
\end{IEEEImpStatement}


%
\IEEEpeerreviewmaketitle

\section{Introduction}

\IEEEPARstart{A}{s} a flexible three-dimensional (3D) data format, point cloud has been widely used in numerous vision applications, including autonomous driving, robotics, medical treatment, etc. The point cloud is a collection of unconstrained points that effectively represents objects and scenes, which are typically obtained by applying scanning or sampling techniques to the surfaces of the target 3D shape. Each point within the point cloud comprises a coordinate tuple $(x,y,z)$, which is treated as the geometry information, and additional attribute information such as color, reflectance, and normal. {\color{black}With the ongoing progress of point cloud acquisition technology~\cite{zhang2024nerf,volumetric_human} and the fast development of 3D vision applications~\cite{fang2023head,survey_detection,PCC_Standard}, the efficient analysis of point cloud shapes has emerged as one of the key focuses in both industry and academia~\cite{survey_autodriving,pc_survey,completion_survey}.}

\subsection{Background}

{\color{black}To address the challenge of point cloud analysis task, which lies in tackling the unordered and irregular points, the community has developed three categories of techniques: multi-view-based~\cite{lawin2017deep,yang2019learning,hamdi2021mvtn,ning2024dilf}, voxel-based~\cite{qi2016volumetric,meng2019vv,zhang2022pvt,OctFormer}, and point-based methods~\cite{qi2017pointnet,wang2019dynamic,zhao2021point,li2023exploiting,liang2024pointmamba}.} The multi-view-based methods~\cite{lawin2017deep,yang2019learning,hamdi2021mvtn,ning2024dilf} first project the point cloud into 2D images, and then  capitalize on conventional 2D techniques to analysis the point cloud shape. The voxel-based methods~\cite{qi2016volumetric,meng2019vv,zhang2022pvt,OctFormer} quantize the input point cloud into a grid-based representation, and utilize 3D convolution or sparse convolution~\cite{sparseConv} to conduct feature extraction. However, both 2D projection and voxelization destroy the details of the original point cloud, resulting in information losses and performance bottlenecks. In the past few years, point-based methods, represent by PointNet~\cite{qi2017pointnet,qi2017pointnet++}, DGCNN~\cite{wang2019dynamic}, and Point Transformer~\cite{zhao2021point,PTv2}, have rapidly garnered extensive attention due to its exceptional capacity in directly processing raw points.

{\color{black}Despite the significant performance advancements of the point-based models, there exists various deficiencies in existing approaches, one of the key issues arises from inefficient input embedding for input point coordinates. Previous works usually follow a bottom-up pipeline that starts from the details and progressively down-samples the skeleton~\cite{qi2017pointnet++,OctFormer,PTv2}. The Multi-Layer Perceptron (MLP)~\cite{qi2017pointnet} is used to simply map the original 3D coordinates into higher dimensions at the initial layer without considering global information.} However, we argue that integrating global information in the embedding step facilitates the network learning, since it can serve as a guidance to the subsequent local feature aggregations.

Another issue lies in the limited efficiency of neighboring aggregations. Recent research works~\cite{hu2020randla,te2018rgcnn,zhou2021adaptive,guo2021pct,han2022blnet} typically conduct multi-scale feature aggregation based on built local graphs. However, these local graphs are either constructed by applying K-Nearest Neighboring (KNN) query on 3D spatial domain or high-dimensional feature domain, neglecting the complementary character of different domains. The KNN graph based on spatial domain gathers the points that are spatially related to each other, and the feature-domain graph links points with long-distance semantic relations. Considering the neighboring points at dual domain is able to enhance the neural model's ability to appreciate both local geometric relations and long-distance semantic connections, which better improves the efficiency of feature aggregations.

\subsection{Our Approach}

To address the above-mentioned issues, we propose a Global Attention-guided Dual-domain feature learning network (GAD). To be specific, we first design a Contextual Position-enhanced Transformer (CPT) module to fully exploits the prior knowledge of the input point cloud shapes, to produce a global-aware input embedding which serves as a guidance to the subsequent aggregations. The devised CPT module is armed with an improved global attention mechanism to best characterize the point cloud shape from the raw input points. Then, the Double K-nearest neighbor Feature Fusion (DKFF) is cascaded to provide efficient feature aggregation by extracting and fusing local graph dynamics that are obtained in both spatial and feature domains. The feature domain focuses on interacting with points that are semantically related without the limitation of distance constraints, while the spatial domain focuses on the points that are spatially related in 3D coordinate space, which facilitates the feature learning for local geometric details. A ResNet-like~\cite{he2016deep} fashion that considers skip connections crossing the network stem (i.e., the stacked DKFF modules) and branches (i.e., specific modules for classification and segmentation tasks) is adopted in our pipeline to facilitate network learning. We examine the efficiency of the proposed model on multiple point cloud analysis tasks as well as a variety of datasets, including classification on ModelNet40~\cite{wu20153d} and ScanObjectNN~\cite{uy2019revisiting}, part segmentation on ShapeNetPart~\cite{yi2016scalable}, and indoor scene segmentation on S3DIS~\cite{armeni20163d}.

The main contributions of this paper can be summarized as:

\begin{itemize}
    \item We propose the Contextual Position-enhance Transformer (CPT) module which is armed with an improved global attention mechanism, to produce a global-aware input embedding that serves as the guidance to subsequent aggregations and facilitates network learning.
    
    \item We propose the Double K-nearest neighbor Feature Fusion (DKFF) module that provides highly effective feature aggregation by conducting novel dual-domain feature learning. It enhances the neural model's ability to appreciate both local geometric relations and long-distance semantic connections.

    \item Levering the proposed CPT module for effective global-aware input embedding and DKFF module for dual-domain feature aggregation, our method attains state-of-the-art performance on multiple point cloud analysis tasks (e.g., classification, part segmentation, and semantic segmentation).  
\end{itemize}

After reviewing related work in Sec. II, we elaborate the proposed method in Sec. III. Experiments are provided in Sec. IV and the conclusion is drawn in Sec. V.

\section{Related work}

\subsection{Multi-view based Methods}
The multi-view based methods first project 3D point clouds into multiple 2D planes, then use 2D image feature extraction and fusion techniques to analysis point cloud shapes. How to aggregate multiple visual features into a discriminative global feature representation is the key challenge. MVCNN \cite{su2015multi} is a pioneering work that maximizes the features of multiple views into a global descriptor, but there is a loss of non-maximum element information. MHBN \cite{yu2018multi} integrates local features via coordinated bilinear pooling. In addition, Yang \emph{et al.} \cite{yang2019learning} utilize relational networks to mine interrelationships on a set of views and then aggregate them to obtain an overall object representation. View-GCN \cite{wei2020view} uses multiple views as graph nodes, and applies local graph convolution, non-local message passing, and selective view sampling to the constructed graph to form a global shape descriptor. {However, the transition to a multi-view representation inevitably results in significant loss of original information from the input point cloud data.}

\subsection{Voxel-based Methods}
Voxel-based methods usually voxelize the point cloud into 3D grids, and then apply 3D convolutions on the volumetric representation for feature extraction. Wu \emph{et al.} \cite{wu20153d} proposed a convolutional deep belief-based 3D ShapeNet to learn the distribution of points in various 3D shapes. But it does not scale well to dense datasets. To this end, OctNet \cite{riegler2017octnet} first uses a hybrid grid-octree structure to divide the point cloud to reduce computational costs. Wang \emph{et al.} \cite{wang2017cnn} proposed an Octree-based CNN to send the average normal vector of the sampling model in the finest leaf octagon to the network to achieve shape classification. PointGrid \cite{le2018pointgrid} combines point and grid representations, enabling the network to extract geometric details. In addition, Ben-Shabat \emph{et al.} \cite{ben20173d} used the 3D modified Fisher Vector (3DmFV) method to represent the three-dimensional grid, and then used the traditional CNN architecture to learn the global representation. Generally speaking, { voxel-based models is constrained by the distortion introduced by voxelization step, leading to performance bottlenecks.}

\subsection{Point-based Methods}
{Point-based methods directly operate on raw point coordinates without additional preprocessing steps, which can be roughly divided into four representative types: point-based MLP, convolution-based, graph-based and attention-based methods.}

\subsubsection{Point-based MLP Methods}Such methods mainly use multiple shared multi-layer perceptrons to model each point independently. As a pioneering work, PointNet \cite{qi2017pointnet} applies a shared multi-layer perceptron to each independent point, and then uses a symmetric aggregation function to aggregate global features, which perfectly adapts to the disorder of the point cloud but ignores the connection with surrounding points. Subsequent PointNet++ \cite{qi2017pointnet++} uses PointNet hierarchically to capture fine geometric structures from the neighborhood of each point. In addition, Duan \emph{et al.} \cite{duan2019structural} proposed a Structural Relational Network(SRN) using MLP to learn structural relational features between different parts.

\subsubsection{Convolution-based Methods}Unlike fixed convolution kernels for 2D images, convolution kernels for 3D point clouds are difficult to design due to the irregularity of point clouds. RSCNN \cite{liu2019relation} implements convolution by learning the mapping from low-level relations such as Euclidean distance and relative position between points in the local subset to high-level relations. In addition, Thomas \emph{et al.} \cite{thomas2019kpconv} use a set of learnable kernels as point cloud rigid and deformable Kernel Point Convolution (KPConv) operators. Whereas in PointConv \cite{wu2019pointconv}, a clip is defined as a Monte Carlo estimation of a continuously sampled 3D convolution with a convolution kernel consisting of a weighting function and a density function. In addition, PCNN \cite{atzmon2018point} also proposed a 3D point cloud convolution network based on radial basis function.

\subsubsection{Graph-based Methods}{Graph-based methods treat each point of the point cloud as a vertex of a graph, and generate directed edges of the graph from neighbors.} ECC \cite{simonovsky2017dynamic} utilizes filter generation network and maximum pooling to aggregate domain information. However, DGCNN \cite{wang2019dynamic} constructs graphs in feature space and dynamically updates between layers. KCNet \cite{shen2018mining} learns relevant features based on the kernel and calculates the affinity between the kernel and the neighborhood of a given point. There are also methods that define convolution as spectral filtering, implemented by multiplying the signal on the graph with the eigenvectors of the Laplacian matrix. For example, RGCNN \cite{te2018rgcnn} treats the entire point cloud as a complete graph and updates the Laplacian matrix at each layer. In PointGCN \cite{zhang2018graph}, KNN is used to find neighbors and then Gaussian kernel is used to weight each edge. The convolution filter is defined as a Chebyshev polynomial in the graph domain, and features are captured by pooling.

\subsubsection{Attention-based Methods}With the success of Transformer in the field of natural language processing, it has also been introduced into 3D vision tasks and has been continuously improved. The attention mechanism is mainly to {generate and allocate the weights to features or neighboring points.} AdaptConv \cite{zhou2021adaptive} uses the attention mechanism to design an adaptive graph convolution kernel to calculate and distinguish the different contributions of neighboring points to the center point. RandLA-Net \cite{hu2020randla} uses Cartesian coordinates and point feature splicing to learn spatial weights to complete local feature aggregation. PointANSL \cite{yan2020pointasnl} utilizes adaptive sampling to propose a local-nonlocal module to capture the local and long-range dependencies of sampling points. PCT \cite{guo2021pct} adopts the same architecture as PointNet \cite{qi2017pointnet}, and proposes Offset-Attention to improve the traditional self-attention. And PT \cite{zhao2021point} does subtraction between query and key to get the channel attention score of the vector, which greatly improves the performance. In addition, PVT \cite{zhang2021pvt} deeply combines the advantages of point-based and voxel-based networks into Transformer, and proposes a local attention module that attains high efficiency and low computational overhead.

\subsection{Summary}

{\color{black}

In summary, multi-view based methods~\cite{su2015multi,yu2018multi,yang2019learning,wei2020view} may suffer from information loss during projection, and voxel-based methods~\cite{wu20153d,riegler2017octnet,wang2017cnn,le2018pointgrid,ben20173d} face challenges with distortion and scalability. Point-based methods exhibit high flexibility and fidelity by directly operating on raw data, and showcasing advancements equipped with MLP~\cite{qi2017pointnet,qi2017pointnet++,duan2019structural}, convolution~\cite{liu2019relation,thomas2019kpconv,wu2019pointconv,atzmon2018point}, graph~\cite{simonovsky2017dynamic,wang2019dynamic,shen2018mining,te2018rgcnn,zhang2018graph}, and attention~\cite{zhou2021adaptive,hu2020randla,yan2020pointasnl,guo2021pct,qi2017pointnet,zhao2021point,zhang2021pvt} mechanisms. Recent attention-based point models, in particular, have shown promising results in 3D vision tasks by effectively capturing local and global dependencies, without any information loss caused by projection and voxelization distortion.

Compared to the prior point-based methods in the field, the key improvements and novelties of the proposed GAD can be highlighted as follows:

\begin{itemize}
    \item Global-aware input embedding. The CPT module incorporates global information into the initial input embedding, providing a more informative representation that guides subsequent local feature aggregations.
    \item Improved global attention mechanism. The CPT employs an enhanced attention mechanism to effectively characterize the overall point cloud shape from raw input points, capturing global context.
    \item Dual-domain feature learning. The DKFF module performs feature aggregation using both spatial and feature domain local graphs, capturing local geometric relations and long-distance semantic connections simultaneously.
    \item The combination of CPT and DKFF modules enables GAD to achieve superior performance on multiple point cloud analysis tasks, including classification, part segmentation, and semantic segmentation.
\end{itemize}
}

\section{Methodology}
We propose a global attention-guided dual-domain feature learning network for point cloud classification and segmentation tasks. This section first outlines the overall network structure and then elaborates the technical details of each module.

\begin{figure*}[t]
    \centering
    \includegraphics[width=1.0\linewidth]{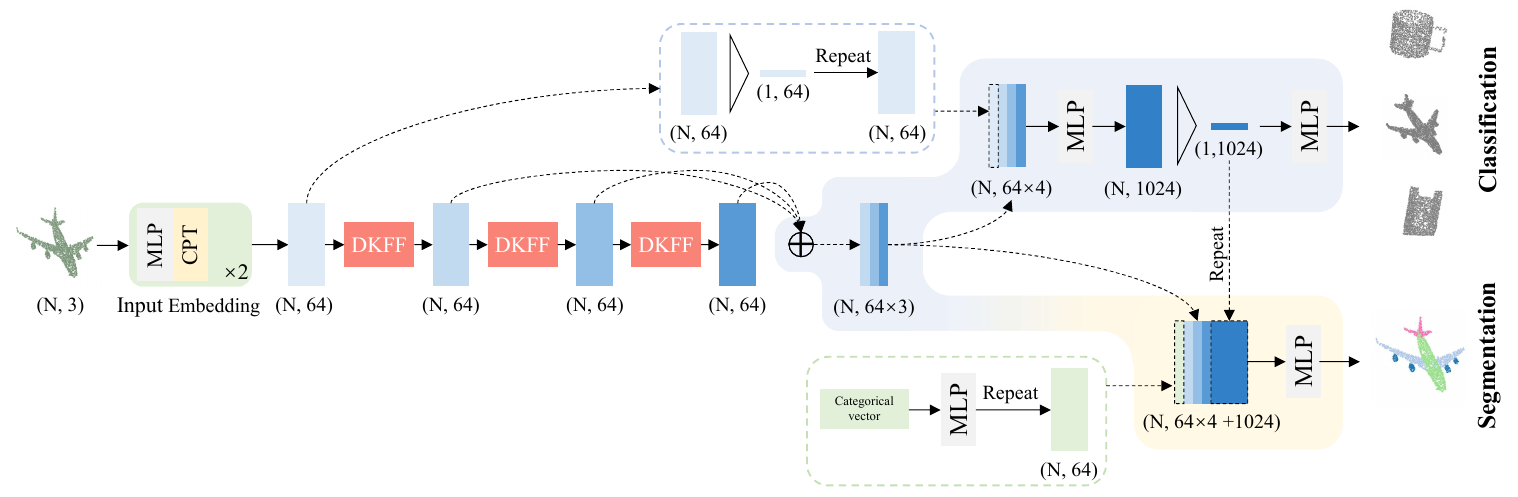}
    \caption{Network architecture of proposed method. ``MLP'' refers to the Multi-Layer Perceptron; ``CPT'' represents the devised Contextual Position-enhanced Transformer module; ``DKFF'' means the Double K-nearest neighbor Feature Fusion module; $N$ refers to the number of points of the input point cloud.}
\label{architecture}
\end{figure*}

\subsection{Framework}

The network architecture of our proposed method is shown in Fig. \ref{architecture}. We initially characterizes the general geometric features of the input point cloud in the shared stem of the network, then two downstream branches are employed for specific tasks (e.g., classification and segmentation). For ease of expression, we use the term ``\emph{stem}'' to denote the common feature extraction phase preceding the specialized task branches.

\subsubsection{Shared Network Stem} As shown in Fig.~\ref{architecture}, we first introduce Contextual Position-enhanced Transformer (CPT) to embed the raw point coordinates into feature space, leveraging a global attention mechanism that efficiently exploits the information of the full point cloud shapes. Based on this, Double K-nearest Feature Fusion (DKFF) modules are cascaded to enhance the obtained global features with locality-aware aggregation in both spatial and feature domains. In this way, the provision of global information from the CPT module serves as guidance for the subsequent dual-domain feature learning, facilitates effective geometric modeling for both global shapes and local details.

\subsubsection{Classification Branch} The features produced by the CPT module incorporating global shape information are aggregated through max-pooling to form a one-dimensional global vector for the initial provision of the global shape context. Then, multi-level features derived from DKFF modules are concatenated and utilized to further enhance and modulate the global vector with additional details.

\subsubsection{Segmentation Branch} Similar to the classification branch, the segmentation branch embraces the integration of global and local features, yet it incorporates a deeper global feature, i.e., the ultimate vector that precedes the classification head in the classification task. In addition, category vector is also concatenated into features as additional messages to the neural model. The concatenated feature, which best describes the geometric information of the input sample, are followed by a multi-layer perceptron that specifies a segmentation label for each point in the point cloud.


\subsection{Contextual Position-Enhanced Transformer (CPT)}

Position encoding is essential for tasks that involve modeling and analyzing the geometry of a point cloud based on its spatial coordinates. Previous works typically utilized a multi-layer perceptron to map three-dimensional coordinates into higher dimensions as the initial features. However, simple mapping techniques do not enable the network to grasp the information of the point cloud geometry due to insufficient interaction within the point set, leading to ineffective feature aggregations. To address this issue, we develop the Contextual Position-enhanced Transformer (CPT) module to generate effective input embedding with integrated global information that severs as a priori guidance to subsequent feature aggregations.

\begin{figure*}[t]
    \centering
    \includegraphics[width=0.8\linewidth]{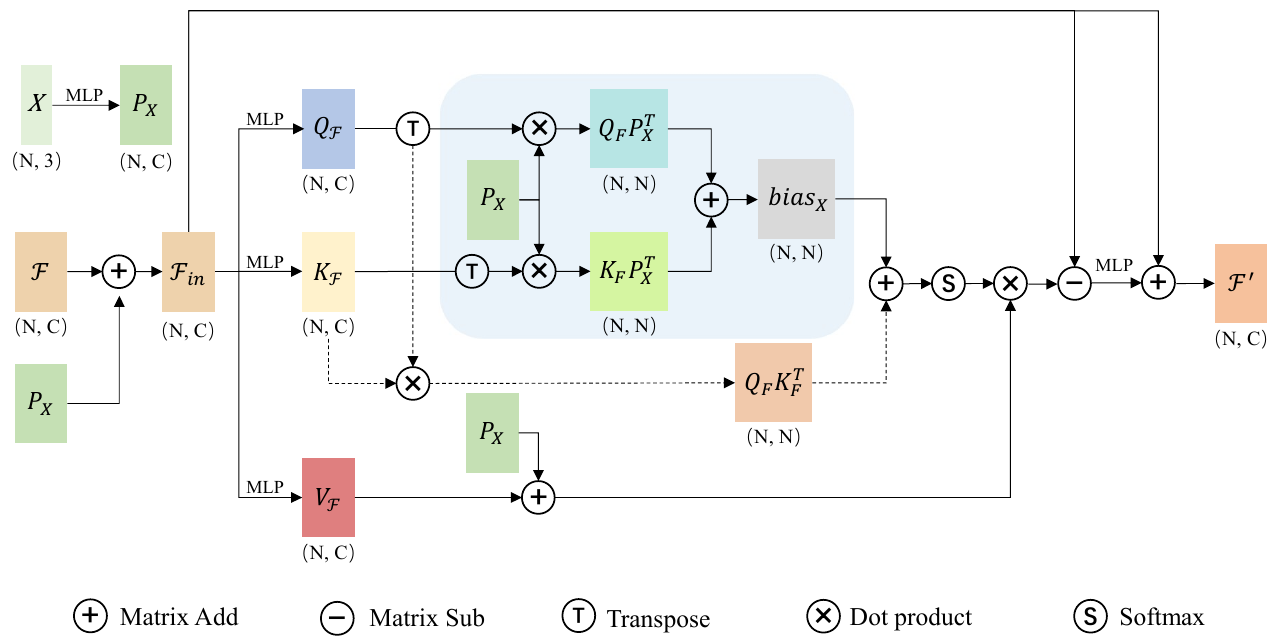}
    \caption{ Proposed Contextual Position-enhanced Transformer (CPT) module. $X$ refers to the original point cloud coordinates; $F$ refers to the point cloud features; $N$ represents the number of points of the input point cloud; $C$ denotes the dimension of feature channel; MLP means multilayer perceptron.}
    \label{CPT}
\end{figure*}

The detailed structure of devised CPT module is shown in Fig. \ref{CPT}. To be specific, we first define the original input point cloud as $X=\left\{x_i|i = 1,2,...,N\right\}\in\mathbb{R}^{N\times3}$, where $x_i$ represents the three-dimensional coordinates $(x,y,z)$ of the $i$-th point. We use the raw coordinates to obtain the naive position embedding $P_{X}$ through the shared MLP, which can be expressed as:
\begin{equation}
P_{X} = MLP (X)
\end{equation}

Then, the naive position embedding $P_{X}$ is added with the point cloud feature $F$ to produce the input feature $F_{in} \in \mathbb{R}^{N \times C}$ for the subsequent attention mechanism. Note that the feature $F$ of the first CPT module is initialized by an non-linear mapping of the input points, e.g., $F=MLP(X)$.

{\color{black}Next, we devise an attention mechanism that is based on contextual position bias to effectively calculate the semantic similarity among all points in the input point cloud.} Different from the previous attention mechanism that directly use the relation calculated by the \emph{query} and \emph{key} as the position bias, we consider a deeper interaction between the position embedding $P_X$, \emph{query}, and \emph{key} to get the position bias with abundant contextual information. Specifically, let $Q_{F}$, $K_{F}$, and $V_{F}$ be the \emph{query}, \emph{key}, and \emph{value} matrices that generated by linear transformation on the feature $F_{in}$ respectively, we calculate the contextual position bias ${bias}_X$ as follows:
\begin{equation}
    {bias}_X = Q_F \times P_X^T + K_F \times P_X^T
\end{equation}

Based on this, the employed attention mechanism can be expressed as follows:
\begin{equation}
    F_{sa} = SoftMax(\frac{Q_F \times K_F^T + bias_X}{\sqrt{C}}) \times (V_F + P_X)
\end{equation}
where $C$ represents the feature channel dimension; $F_{sa}$ refers to the feature by the attention mechanism.

{\color{black}Given that $Q_F$, $K_F$, and $V_F$ are derived from the high-dimensional features, employing a conventional attention mechanism directly in this feature space is empirically inefficient, due to the risk of an excessive focus on high-level semantic information and neglect of the shape details. The contextual position bias $bias_X$ serves as a timely injection of raw point cloud geometry position, effectively supplements detailed information in high-dimensional semantic features.}

As the final step, the CPT module adpots the offset calculation between the self-attention features and the input features by subtraction, similar to the scheme used in PCT~\cite{guo2021pct}, to obtain better network performance. Mathematically, 
\begin{equation}
F^{'} = MLP(F_{in} - F_{sa}) +F_{in}
\end{equation}
where $F^{'}$ represents the output of the proposed CPT module.

\subsection{Double K-nearest neighbor Feature Fusion (DKFF)} 

\begin{figure*}[t]
    \centering
    \includegraphics[width=0.9\linewidth]{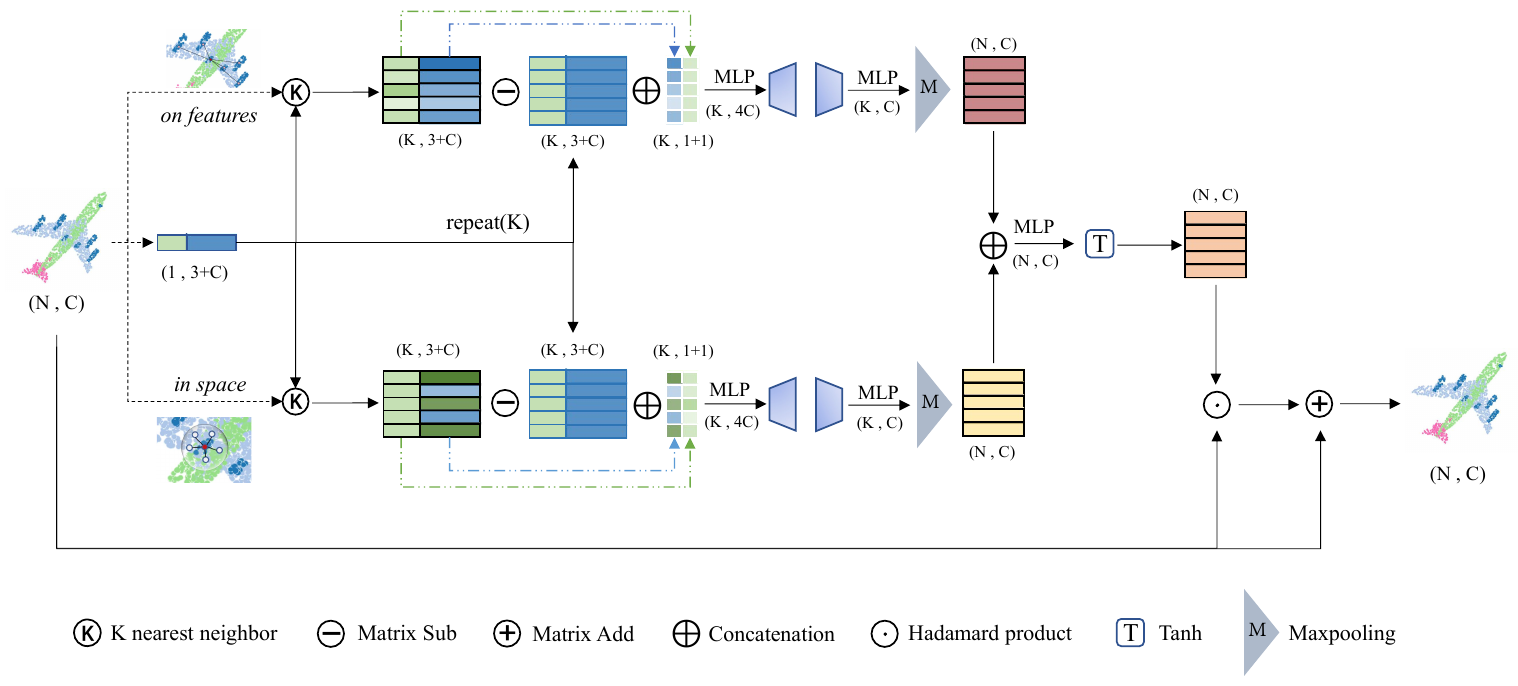}
    \caption{Proposed Double K-nearest neighbor Feature Fusion (DKFF) module. $N$ represents the number of points of the input point cloud; $C$ denotes the dimension of feature channel; MLP means multilayer perceptron.}
    \label{DKFF}
\end{figure*}

The fashion of directly constructing local graphs in the feature domain has been proven to be effective for point cloud analysis tasks, owing to the ability to capture potentially long-distance semantic characteristics~\cite{wang2019dynamic}. However, the neighbor querying in feature domain almost completely ignores the geometric correlation in the original coordinate domain. {\color{black} For instance, in the shape of airplane, the points on the left engine may be aggregated with points on the right engine after feature domain queries, but the local geometry relationship between the left engine and wings may be ignored.} To address this issue, we propose a Double K-nearest neighbor Feature Fusion (DKFF) for complementary fusion learning in both the coordinate and the feature domain.

Specifically, for each point in the point set, we use K-Nearest Neighbor (KNN) to construct local graphs in the spatial domain (i.e., the coordinate set $X$) and feature domain (i.e., the point features $F$). Let $x_i$ and $f_j$ denotes the coordinate and corresponding feature of the $i$th point, respectively, then this process can be represented as:
\begin{equation}
    \left \{ x^j_i \right \}_{j=1}^{K} = KNN(x_{i},X), \quad \left \{ f^k_i \right \}_{k=1}^{K} = KNN(f_{i},F)
\end{equation}

Then, numerous critical elements in the local graph are identified and gathered to effectively aggregate features for $i$th point. Note that the aggregation in this stage is independently conducted for each spatial and feature domain. Take spatial domain as an example, we aggregate the direction vector $(x_i-x_i^j)$, feature subtraction offset $(f_i-f_i^j)$, inter-point distance $\parallel x_i-x_i^j \parallel_{2}$, and averaged feature distance $L_i^j$ to characterize geometry dynamics $p_{i}$ of the obtained spatial graph:
\begin{equation}
    p_{i} = MLP \left\langle(x_i-x_i^j), (f_i-f_i^j), \parallel x_i-x_i^j\parallel_{2},L_i^j \right\rangle
\end{equation}
\begin{equation}
    L_i^j = \frac{1}{C} \parallel f_i-f_i^j \parallel_{1}
\end{equation}
where $\parallel \cdot \parallel_{2}$ represents the Euclidean distance between points, $\left \langle \cdot \right \rangle$ represents the concatenate operation, $\parallel \cdot \parallel_{1}$ represents the L1 norm, $C$ refers to the channel dimension. Correspondingly, the feature dynamics $q_{i}$ of the graph in the feature domain is expressed as:
\begin{equation}
    q_{i} = MLP \left\langle(x_i-x_i^k), (f_i-f_i^k), \parallel x_i-x_i^k\parallel,L_i^k \right\rangle
\end{equation}
\begin{equation}
    L_i^k = \frac{1}{C} \parallel f_i-f_i^k \parallel_{1}
\end{equation}

We leverage the inverse bottleneck design~\cite{qian2022pointnext,sandler2018mobilenetv2} to enrich feature extraction by expanding the output channels of hidden layer by 4 times in the subsequent MLP, as shown in Fig.~\ref{DKFF}. Then, maxpooling operation $Max$ is used to aggregate the geometry features of each KNN graph to a feature vector:
\begin{equation}
    P_i = Max (MLP(MLP(p_{i})))
\end{equation}
\begin{equation}
    Q_i = Max (MLP(MLP(q_{i})))
\end{equation}
where $P_{i} \in \mathbb{R}^{1 \times C}$ refers to the aggregated feature in the spatial domain and $Q_i \in \mathbb{R}^{1 \times C}$ represents the feature in the feature domain. Here, $P_{i}$ focuses on capturing geometry dynamics from spatial neighborhoods, whereas $Q_{i}$ appreciates the feature-level neighbors that transcends spatial limitations, effectively complementing $P_{i}$.

To facilitate information exchange across different domains, we concatenate the aggregated feature of both domain, followed by a multilayer perceptron and a $Tanh$ function to provide a residual multiplier $W_{i} \in \mathbb{R}^{1 \times C}$:
\begin{equation}
    W_{i} = Tanh(MLP\left\langle P_i,Q_i \right \rangle)
\end{equation}
where $\left \langle \cdot \right \rangle$ represents the concatenate operation. The $Tanh$ function, which manifests symmetricity around the origin, restricting the obtained multiplier within the range of [-1,1] to better serves for the subsequent multiplication. Let $W \in \mathbb{R}^{N \times C}$ be the learned weight matrix that contains $N$ residual multipliers (each multiplier corresponds to a point), we use Hadamard product to weight the input features by $W$, followed by a residual connection~\cite{he2016deep} to produce the output feature:
\begin{equation}
    F_{out} = F_{in} \cdot W +F_{in}
\end{equation}

The way of enhancing features through dual-domain aggregation and residual connections in the proposed DKFF module expands the receptive field of the neural network and mitigates the problem of gradient vanishing in deeply cascaded network modules. Beyond that, semantic information can be effectively exploited through interactive learning across different domains, which enhances the ability for point cloud understanding of the neural model.

\section{Experiments} 

In section, we evaluate our method on multiple point cloud analysis tasks including classification, part segmentation, and indoor large-scale semantic segmentation. State-of-the-art methods are compared for each task and extensive ablation studies are conducted to examine the effectiveness of our network structure.

\subsection{Classification}
{\bf Data. }We use the ModelNet40  \cite{wu20153d} and ScanObjectNN  \cite{uy2019revisiting} datasets for point cloud classification evaluation. ModelNet40  \cite{wu20153d} contains {12,311} meshed CAD models from 40 categories, of which {9,843} models are used for training and {2,468} models are used for testing. We follow the experimental setting of  \cite{qi2017pointnet} and uniformly sample each object into {1,024} points containing only 3D coordinates as input. ScanObjectNN  \cite{uy2019revisiting} contains about {15,000} real scanned objects {which are further} grouped into 15 classes with {2,902} unique object instances. {Since point cloud samples in ScanObjectNN dataset are scanned from the real world, there exists} noise caused by occlusion and missing background, which poses a major challenge to the existing point cloud analysis methods. It is also sampled into {1,024} points containing only three-dimensional coordinates as input.

{\bf Network configuration.} {Python and Pytorch is used to implement our model, and all experiments are conducted on two RTX 2080Ti GPUs.} On ModelNet40 \cite{wu20153d}, the neighbor k value is selected as 16. Due to the noise in the ScanObjectNN  \cite{uy2019revisiting} data, the k value is selected as 20, which is slightly larger. All layers use LeakyReLU and batch normalization. We use the SGD optimizer with momentum set to 0.9. The initial learning rate is 0.1 and reduced to 0.001 using cosine annealing. The batch size is set to 32. To prevent the network from overfitting, the random drop rate is {set to} 0.5, and the training is performed for 200 epochs. The data augmentation process all includes point displacement, scaling and perturbation.

{\bf Results.} {The classification results on ModelNet40  \cite{wu20153d} are shown in Tab. \ref{classification_M} with the evaluation metrics of mean class accuracy (mAcc) and overall accuracy (OA). The input data format and number of points are also provided for a detailed comparison.} It can be seen that our method achieves the best {performance on overall accuracy} using only 1k points containing 3D coordinates, which is significantly better than other methods. {\color{black}Note that the latest published works are also included, such as M-GCN \cite{hu2023m}, PointConT  \cite{liu2023point}, IBT \cite{li2023exploiting}, OctFormer~\cite{OctFormer}, and PointMamba~\cite{liang2024pointmamba}.}

\begin{table}[h]
\begin{center}
\caption{Classification results on ModelNet40.}
\label{classification_M}
\scalebox{0.85}{\begin{tabular}{l c c c c}
\hline
Methods &Input & point & mAcc & OA\\
\hline
Other Learning-based Methods \\
\hline
Pointnet \cite{qi2017pointnet}&xyz & 1k & 86.0 & 89.2 \\
Pointnet++ \cite{qi2017pointnet++}&xyz,normal & 5k & - & 91.9 \\
PointCNN \cite{li2018pointcnn}&xyz & 1k & 88.1 & 92.2\\
DGCNN \cite{wang2019dynamic}&xyz & 1k & 90.2 & 92.2\\

SpiderCNN \cite{xu2018spidercnn}&xyz,normal & 1k & - & 92.4 \\
PointWeb \cite{zhao2019pointweb}&xyz,normal & 1k & 89.4 & 92.3 \\
PointConv \cite{wu2019pointconv}&xyz,normal & 1k & - &92.5\\
Point2Sequence \cite{liu2019point2sequence}&xyz & 1k & 90.4 & 92.6\\
KPConv \cite{thomas2019kpconv}&xyz & 6k  & - & 92.9\\
FPConv \cite{lin2020fpconv}&xyz,normal & 1k  & - & 92.5\\
Point2Node \cite{han2020point2node}&xyz & 1k & - & 93.0\\
M-GCN \cite{hu2023m}&xyz & 1k & 90.1 &93.1\\
AG-conv \cite{zhou2021adaptive} &xyz & 1k&90.7 &93.4\\
PointStack \cite{wijaya2022advanced}&xyz & 1k & 89.6 & 93.3\\
\textcolor{black}{PointMamba} \cite{liang2024pointmamba} & xyz & 1k & - & 92.4 \\
\hline
Transformer-based Methods \\
\hline
A-SCN \cite{xie2018attentional}&xyz & 1k & 87.6 &90.0\\
PATs \cite{yang2019modeling}&xyz & 1k & - &91.7\\
GAPNet \cite{chen2019gapnet}&xyz,normal & 1k & 89.7 &92.4\\
LFT-Net \cite{gao2022lft}&xyz,normal & 2k & 89.7 &93.2\\
3DETR \cite{misra2021end}&xyz & 1k & 89.9 &91.9\\
MLMST \cite{zhong2021point}&xyz & 1k &- &92.9\\
PCT \cite{guo2021pct}&xyz & 1k & - &93.2\\
CloudTransformers \cite{mazur2021cloud}&xyz & 1k & 90.8 &93.1\\
3DCTN \cite{xie2018attentional}&xyz,normal & 1k & 91.2 &93.3\\
PointASNL \cite{yan2020pointasnl}&xyz & 1k & - &92.9\\
PointASNL \cite{yan2020pointasnl}&xyz,normal & 1k & - &93.2\\
PT \cite{zhao2021point}&xyz,normal & 1k & 90.6 &93.7\\
PointConT \cite{liu2023point}&xyz & 1k & - &93.5\\
IBT \cite{li2023exploiting}&xyz & 1k & 91.0 &93.6\\
\textcolor{black}{OctFormer} \cite{OctFormer} & xyz & 1k & - & 92.7 \\
\hline
Ours &xyz & 1k& 91.1 & \textbf{93.8}\\
\hline
\end{tabular}}
\end{center}
\end{table}

{Tab. \ref{classification_S} shows the classification results on ScanObjectNN dataset~\cite{uy2019revisiting}, where our method continues to provide superior classification performance.} Due to the defect of the objects within this dataset such as occlusion and noise, this also proves {the significant stability and robustness of our method.}

\begin{table}[h]
\begin{center}
\caption{Classification results on ScanObjectNN.}
\label{classification_S}
\scalebox{0.85}{\begin{tabular}{l c c}
\hline
Methods& mAcc & OA\\
\hline
3DmFV \cite{ben20183dmfv} & 58.1&63.0\\
Pointnet \cite{qi2017pointnet}& 63.4 & 68.2 \\
Spidercnn \cite{xu2018spidercnn}& 69.8 & 73.7 \\
Pointnet++ \cite{qi2017pointnet++}& 75.4 & 77.9 \\
DGCNN \cite{wang2019dynamic}& 73.6 & 78.1 \\
PointCNN \cite{li2018pointcnn}& 75.1 & 78.5 \\
BGA-DGCNN \cite{uy2019revisiting}& 75.7 & 79.7 \\
BGA-PN++ \cite{uy2019revisiting}& 77.5 & 80.2 \\
DRNet \cite{qiu2021dense}& 78.0 & 80.3 \\
GBNet \cite{qiu2021geometric}& 77.8 & 80.5 \\
SimpleView \cite{goyal2021revisiting}&-&80.5\\
PRANet \cite{2021PRA}&79.1&82.1\\
{\color{black}PointMamba \cite{liang2024pointmamba}} & - & 82.5 \\
\hline
Ours & \textbf{80.1} & \textbf{82.6}\\
\hline
\end{tabular}}
\end{center}
\end{table}

\begin{figure*}[t]
    \centering
    \includegraphics[width=16cm,height=10cm]{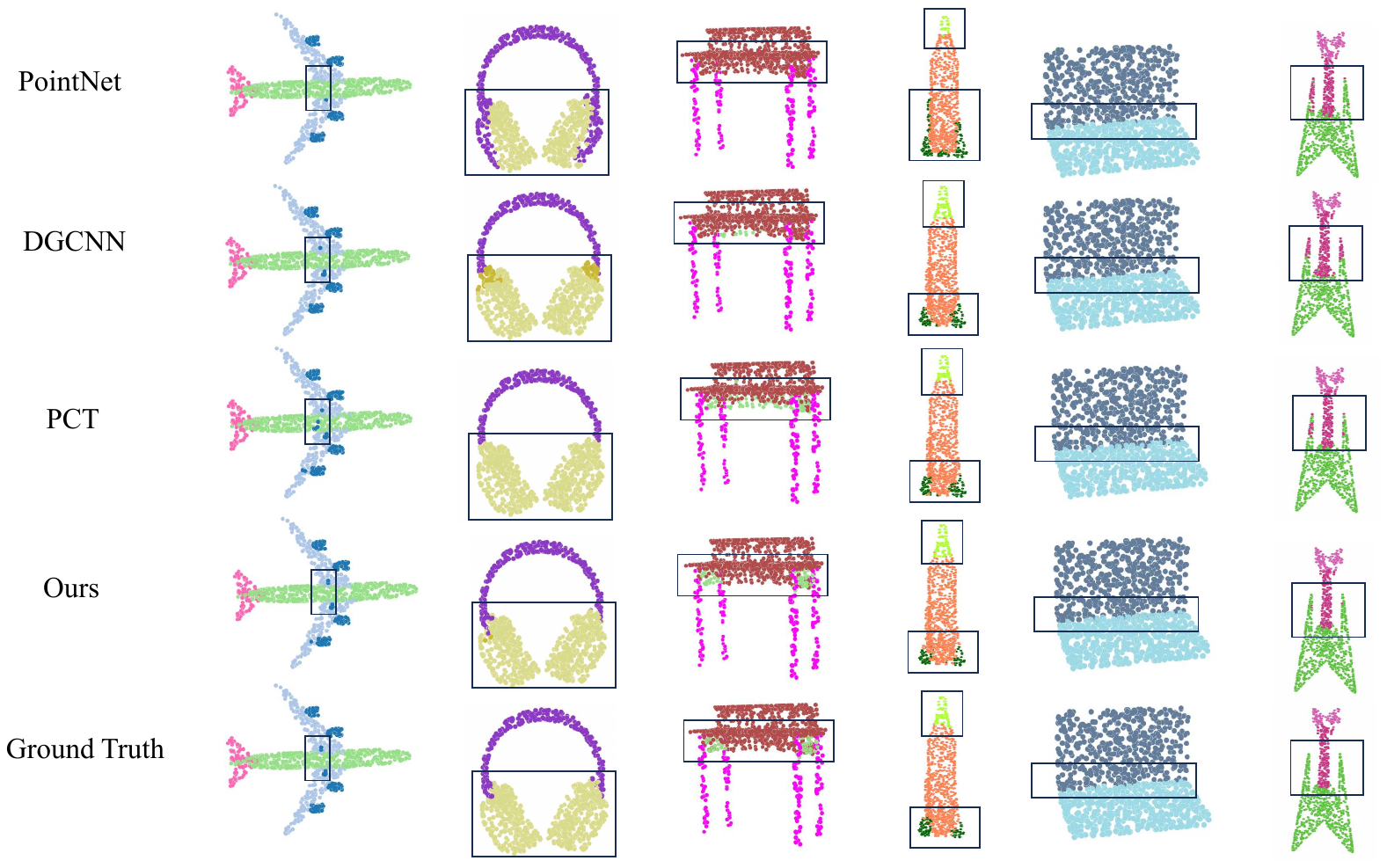}
    \vspace{-5pt}
    \caption{Visual comparison with other methods for part segmentation.}
\label{visual_p}
   \vspace{-10pt}
\end{figure*}

\subsection{Part segmentation}
{\bf Data.} We further test our model on the part segmentation task on the ShapeNetPart  \cite{yi2016scalable} dataset. The dataset contains {16,880} shapes from 16 categories, of which {14,006} are used for training and {2,874} are used for testing. The number of parts in each category ranges from 2 to 6, for a total of 50 different parts. We follow the experimental setup of  \cite{qi2017pointnet++}, but {only sample 1,024} points from each shape instead of {2,048. Our input data include only 3D coordinates} and no point normal is attached.

\begin{table*}[t]
\begin{center}
\caption{Part segmentation results on ShapeNet dataset. Metric is mIoU(\%).}
\label{segmentation}
\resizebox{1\textwidth}{26mm}
{
\begin{tabular}{c|c|c|c|c|c|c|c|c|c|c|c|c|c|c|c|c|c}
\hline
Methods& mIoU & air. & bag & cap  & car & cha. & ear.  & gui. & kni. & lam. & lap. & mot. & mug & pis. & roc. & ska.  & tab.\\
\hline
NUM &  & 2690 & 76 & 55 & 898 & 3758 & 69 & 787 & 392 & 1547 & 451 & 202 & 184 & 283 & 66 & 152 & 5271 \\
\hline
Other Learning-based Methods&  &  &  &   &  &  &   &  &  &  &  &  &  &  &  &   & \\
\hline
Pointnet \cite{qi2017pointnet} & 83.7 & 83.4 &78.7 & 82.5 & 74.9 & 89.6 & 73.0 & 91.5 & 85.9 & 80.8 & 95.3 &65.2 & 93.0 & 81.2 & 57.9 & 72.8 & 80.6 \\
Pointnet++ \cite{qi2017pointnet++} & 85.1 & 82.4 &79.0 & 87.7 & 77.3 & 90.8 & 71.8 & 91.0 & 85.9 & 83.7 & 95.3 &71.6 & 94.1 & 81.3 & 58.7 & 76.4 & 82.6 \\
RGCNN \cite{te2018rgcnn} & 84.3 & 80.2 &82.8 & \textbf{92.6} & 75.3 & 89.2 & 73.7 & 91.3 & 88.4 & 83.3 & 96.0 &63.9 & 95.7 & 60.9 & 44.6 & 72.9 & 80.4\\
\textbf{SO-Net} \cite{li2018so} & 84.9 & 82.8 &77.8 & 88.0 & 77.3 & 90.6 & 73.5 & 90.7 & 83.9 & 82.8 & 94.8 &69.1 & 94.2 & 80.9 & 53.1 & 72.9 & 83.0\\
DGCNN \cite{wang2019dynamic} & 85.2 & 84.0 &83.4 & 86.7 & 77.8 & 90.6 & 74.7 & 91.2 & 87.5 & 82.8 & 95.7 &66.3 & 94.9 & 81.1 & 63.5 & 74.5 & 82.6\\
PCNN \cite{atzmon2018point} & 85.1 & 82.4 &80.1 & 85.5 & 79.5 & 90.8 & 73.2 & 91.3 & 86.0 & 85.0 & 96.7 &73.2 & 94.8 & 83.3 & 51.0 & 75.0 & 81.8\\
3D-GCN \cite{lin2020convolution} & 85.1 & 83.1 & 84.0 & 86.6 & 77.5 & 90.3 & 74.1 & 90.9 & 86.4 & 83.8 & 95.3 &65.2 & 93.0 & 81.2 & 59.6 & 75.7 & 82.8\\
\textcolor{black}{PointMamba \cite{liang2024pointmamba}} & 85.8 & - &- & - & - & - & - & - & - & - &- & - & - & - & - & - &-\\
\hline
Transformer-based Methods&  &  &  &   &  &  &   &  &  &  &  &  &  &  &  &   & \\
\hline
\textbf{A-SCN} \cite{xie2018attentional} & 84.6 & 83.8 &80.8 & 83.5 & 79.3 & 90.5 & 69.8 & 91.7 & 86.5 & 82.9 & 96.0 &69.2 & 93.8 & 82.5 & 62.9 & 74.4 & 80.8\\
\textbf{GAPNet} \cite{chen2019gapnet}  & 84.7 & 84.2 &84.1 & 88.8 & 78.1 & 90.7  & 70.1 & 91.0   & 87.3 & 83.1 & 96.2  & 65.9  & 95.0 & 81.7   &60.7 & 74.9 & 80.8  \\ 
MLMST \cite{zhong2021point}& 86.0 & 83.6 &\textbf{84.7} & 86.3 & 79.8 & 91.1  & 71.2 & 90.2   & \textbf{88.6} & 84.9 & 95.9  & 72.8  & 94.8 & 83.4   &56.2 & 76.7 & 82.6  \\ 
PointASNL \cite{yan2020pointasnl} & 86.1 & 84.1 &\textbf{84.7} & 87.9 & 79.7 & \textbf{92.2} & 73.7 & 91.0 & 87.2 & 84.2 & 95.8 &74.4 & 95.2 & 81.0 & 63.0 & 76.3 & 83.2\\
PCT \cite{guo2021pct} & 86.4 & 85.0 &82.4 & 89.0 & \textbf{81.2} & 91.9 & 71.5 & 91.3 & 88.1 & \textbf{86.3} & 95.8 &64.6 & \textbf{95.8} & 83.6 & 62.2 & \textbf{77.6} & 83.7\\
{ Point2Vec} \cite{abou2023point2vec} & 86.3 & - &- & - & - & - & - & - & - & - &- & - & - & - & - & - &-\\
{ IBT} \cite{li2023exploiting} & 86.2 & 85.2 &81.4 & 86.1 & 80.1 & 91.5 & \textbf{76.6} & \textbf{91.9} & 87.6 & 84.6 &\textbf{97.1} & 72.9 & 95.4 & 84.3 & \textbf{63.7} & 76.5 &83.9\\
\hline
Ours & 86.3 & \textbf{85.4} &82.0 & 84.0 & 80.8 & 91.2 & 76.5 & \textbf{92.0} & 87.7 & 85.3 &96.2 & \textbf{74.8} & 95.2 & \textbf{84.8} & 63.5 & 75.5 &\textbf{84.0}\\
\hline
\end{tabular}}
\end{center}
\end{table*}

{\bf Network configuration.} Following  \cite{wang2019dynamic}, we include a one-hot vector representing category types for each point. We concatenate the high-dimensional global feature vector obtained in the classification task and categorical vector with the previously learned semantic features of different levels to predict the category of each point. The settings of other training parameters are the same as our classification task, except that the neighbor value k is set to 40 to further expand the receptive field to learn fine-grained features.

{\bf Results.} {Tab. \ref{segmentation} reports the part segmentation results of different methods, where we use the mean class IoU (mcIoU) per class and mean instance IoU (mIoU) across all shapes in all categories as metrics. Note that the} IoU of a shape is computed by averaging the IoU of each part {and the} mIoU is computed by averaging the IoUs of all testing instances. Our method performs better on most categories and also performs well in terms of overall mIoU. In addition, we also have a visual comparison of part segmentation with some other mainstream methods, as shown in Fig. \ref{visual_p}. For parts such as aircraft wing engines, rocket heads and tails, guitar strings, etc., our method is significantly closer to the ground truth.

\begin{figure*}[t]
    \centering
    \includegraphics[width=16cm,height=9cm]{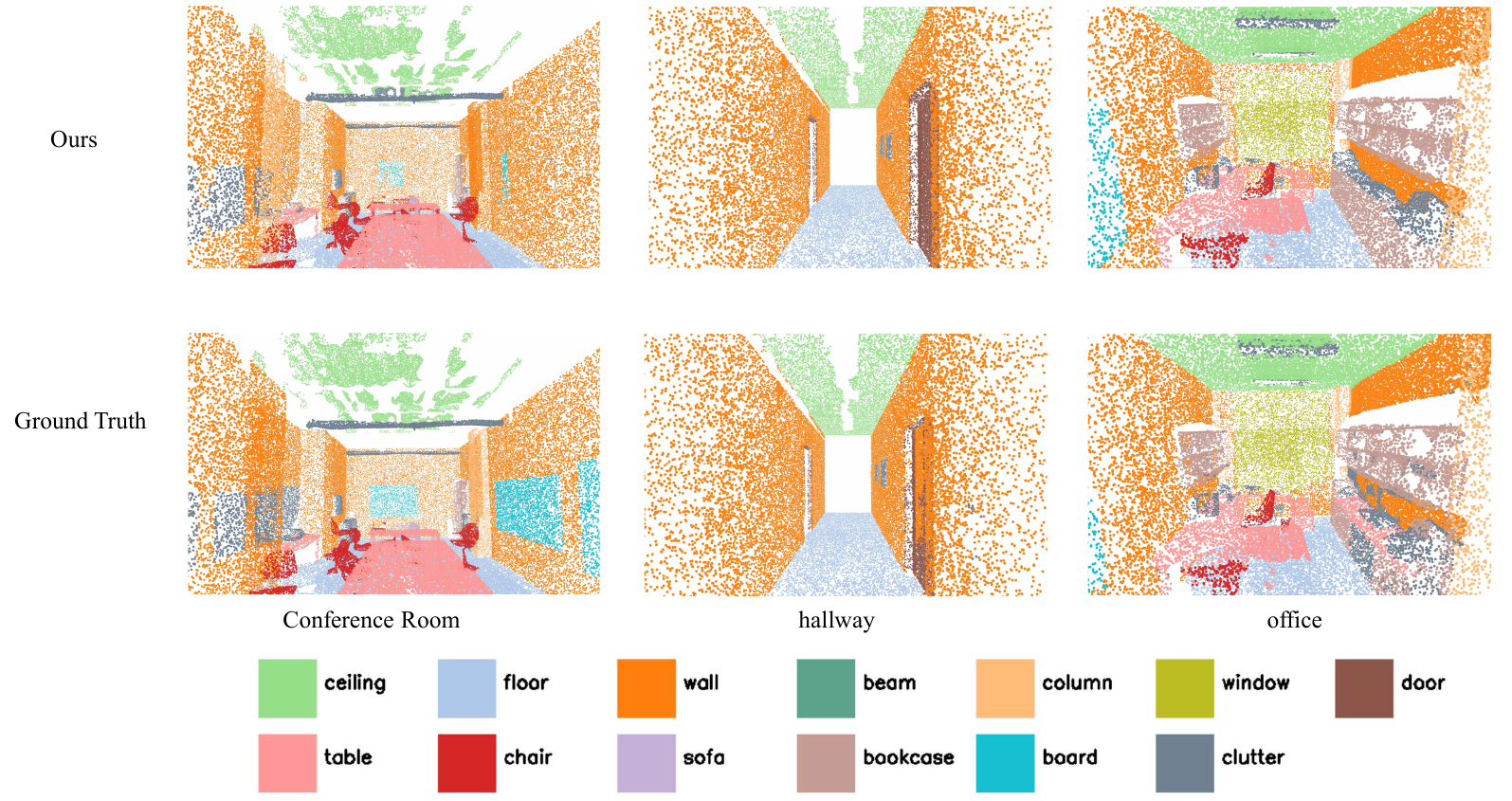}
    \caption{Visualization of semantic segmentation results on the S3DIS dataset.}
\label{visual_s}
\end{figure*}

\subsection{Indoor scene segmentation}
{\bf Data.} We further test the semantic segmentation performance on the large-scale dataset S3DIS  \cite{armeni20163d} {which} contains 3D RGB point clouds of six indoor areas from three different buildings {with} a total of 271 rooms. Each point is annotated with a semantic label from 13 categories. We follow the experimental settings of  \cite{wang2019dynamic} {to divide} the original large-scale point cloud data into {1m × 1m} blocks, {and} randomly sample {4,096} points {within} each block. {Each block is then treated as independent input for the neural network.} The attributes of the input point include {coordinates, color,} and normalized spatial coordinates. We {choose} Area 5 as the test set which is not in the same building as other areas.

{\bf Network configuration.} The training parameters and network structure are the same as the part segmentation, but no additional class vectors are introduced. {Given the increased complexity involved in large-scale semantic segmentation task, we increased the stack depth of the DKFF module to 5 layers} to encourage the network to learn richer semantic features.

{\bf Results.} We report the mIoU, mean classwise accuracy (mAcc) and overall accuracy (OA) in Tab. \ref{segmentation result} { and provide visualization results in Fig. \ref{visual_s}. As seen, our method slightly underperforms the state of the arts but there exists reasons for this issue. The primary factor, we believe, lies in the necessity to reduce the input size to 1m × 1m blocks with merely 4k points, to fit the limited hardware resource. In contrast, previous works~\cite{hu2020randla,zhao2021point,lai2022stratified,zhou2021adaptive} usually input more points even the entire scenes for training due to their ample computational resources.}


\begin{table}[h]
\begin{center}
\caption{Segmentation results on S3DIS.}
\label{segmentation result}
\scalebox{0.85}{\begin{tabular}{l c c c}
\hline
Methods& mAcc &OA& mIoU\\
\hline
Pointnet \cite{qi2017pointnet}& 49.0 &-& 41.1 \\
SEGCloud \cite{tchapmi2017segcloud} & 57.3 &-& 48.9\\
RSNet \cite{huang2018recurrent} & 57.3 &-& 51.9\\
PointCNN \cite{li2018pointcnn} & 63.9&85.9 & 57.3\\
PointWeb \cite{zhao2019pointweb} & 66.6 &86.9& 60.3\\
SPG \cite{landrieu2018large} & 66.5 &86.4& 58.0\\
ELGS \cite{wang2019exploiting} &-&88.4&60.1\\
Grid-GCN \cite{xu2020grid} &87.0&86.9&57.8\\
PCT \cite{guo2021pct} & 67.6 &-& 61.3\\
PointASNL \cite{yan2020pointasnl}&68.5 & 87.7  &62.6\\
\hline
Ours & 68.5 & 88.3&62.9\\
\hline
\end{tabular}}
\end{center}
\vspace{-2mm}
\end{table}

{
\subsection{Ablation experiment} 

In this subsection, we explore additional architectural choices of the network and conduct ablation studies on the proposed modules. 
}

{\bf Efficacy of proposed modules.} We design the Contextual Position-enhanced Transformer (CPT) {to produce a global-aware input embedding that serves as the guidance to subsequent aggregations and facilitates network learning.} This has not been {previously investigated} and { differs from existing} methods that directly map 3D coordinates to higher dimensions to obtain initial features. {To validate the efficacy} of this module, {we conduct ablation study by replacing the CPT module with conventional shared MLP to obtain the initial features.}

{The Double K-nearest neighbor Feature Fusion (DKFF) module provides highly effective feature aggregation by conducting novel dual-domain feature learning. In this subsection, we evaluated the model's performance in single domain, either spatial or feature domain, to examine the efficiency of proposed dual-domain learning.}

Results of this ablation study on 1k points ModelNet40~\cite{wu20153d} are shown in Tab. \ref{ab_module}, where \checkmark means to keep the module, and blank means to remove it. It can be seen that the way of expanding the receptive field with dual domains is efficient, and the best performance can be obtained with the initial features enhanced by CPT.

\begin{table}[h]
\caption{Ablation experiments for the proposed CPT and DKFF modules. Classification task is considered as the benchmark task.}
\label{ab_module}
\centering
\scalebox{0.8}{
\begin{tabular}{c| ccc|cc} 
\hline
Model & CPT & DKFF (in space) & DKFF (in feature) & mAcc & OA  \\ 
\hline
Baseline (PointNet)& & & &86.0&89.2\\
A                         & \Checkmark         &\Checkmark        &             & 90.5  & 93.3   \\
B                         & \Checkmark          &       & \Checkmark           & 90.7  & 93.4   \\
C                         &           & \Checkmark         & \Checkmark           & 90.4  & 93.4   \\
D                         & \Checkmark         & \Checkmark        & \Checkmark          & \textbf{91.1}  & \textbf{93.8}   \\
\hline
\end{tabular}}
\vspace{-2mm}
\end{table}

{\bf Module designing.} {We perform investigations into the specifics of the design of proposed CPT and DKFF modules, the results are shown in Tab. \ref{ab_in}.} Several interesting {phenomenons} can be observed. First, the {aim} of adding Euclidean distance and feature average metric is to learn complex geometric patterns, which is more helpful for original noise-free data sets such as ModelNet40  \cite{wu20153d} and ShapeNetPart  \cite{yi2016scalable}. {But for the noisy samples in ScanObjectNN  \cite{uy2019revisiting}, this scheme appears ineffective since the aforementioned auxiliaries introduces interference information on noisy inputs}. Second, learning from self-attention is relatively more fragile and less {efficient} than offset attention. Third, {the fusion of location information and attention weights performs much better than that of ignoring location information}, which fully proves the importance of location information for subsequent point cloud learning.

\begin{table}[h]
\caption{Ablation experiments for different options within the module.}
\label{ab_in}
\centering
\scalebox{0.8}{
\begin{tabular}{c|c|c |c} 
\hline
Module                               &                       & ModelNet40& ScanObjectNN  \\ 
\hline
\multirow{3}{*}{CPT} & w/o $bias_X$            & 93.2
 & 81.2  \\ 
 \cline{2-4}
                                            & use $P_X$ instead of $bias_X$             & 93.4  & 81.8  \\ 
\cline{2-4}
                                            & use SA instead of offset attention                  & 93.0 & 80.3   \\ 
\hline
\multirow{4}{*}{DKFF}            & w/o distance and feature averaging metric   & 93.6  & \textbf{82.6}  \\ 
\cline{2-4}
                          & use average pooling instead of maxpooling& 93.2 & 81.3   \\ 
\cline{2-4}
                          & use attention pooling instead of maxpooling& 93.2 & 80.9   \\ 
\cline{2-4}
                          & use Sigmoid instead of Tanh& 93.5 & 82.2   \\ 
\hline
both&&\textbf{93.8}&81.8\\
\hline

\end{tabular}}
\end{table}

\textcolor{black}{\bf Number of DKFF layer.} {\color{black} We conduct ablation experiments to verify the impact of the number of cascaded DKFF layers. We retrain our network under four different conditions (i.e., 1$\sim$4 layers) and test it using classification benchmarks. Results of this ablation study on 1k points ModelNet40~\cite{wu20153d} are shown in Tab.~\ref{tab:ablation_n_DKFF}. As seen, increasing the depth of stacked layers generally yields improved performance. However, excessively deep structures (e.g., four or more layers) can hinder network learning due to an increase in the number of parameters that require optimization, potentially leading to suboptimal convergence.
}

\begin{table}[h]
\caption{\color{black} Ablation study for the number of cascaded DKFF layers. Classification task is considered as the benchmark task.}
\label{tab:ablation_n_DKFF}
\centering
\scalebox{0.9}{
\begin{tabular}{c|c c} 
\hline
Number of DKFF layer & mAcc & OA  \\ 
\hline
1   & 90.5  & 93.2  \\
2   & 90.8  & 93.5  \\
3   & \textbf{91.1}  & \textbf{93.8}  \\
4   & 91.0  & 93.6  \\
\hline
\end{tabular}}
\end{table}

{\bf k-nearest neighbor value.} Considering the strong association between the feature aggregation efficiency and receptive field size, which is determined by the value k of the KNN gathering in the DKFF module, we conduct ablation experiments to assess the network performance across different k values. As seen in Tab. \ref{ab_k}, a small k value leads to inadequate information exchange within neighboring points, while a large k value may exert complex context patterns, diminishing prediction accuracy. We consider 16$\sim$20 as appropriate values for sparse point cloud samples.

\begin{table}[h]
\caption{\color{black} Ablation study for different k values of the DKFF module. Overall accuracy (OA) for classification benchmark datasets (i.e., ModelNet40 and ScanObjectNN) is reported.}
\label{ab_k}
\centering
\scalebox{0.9}{
\begin{tabular}{c|c c} 
\hline
{\color{black} k value} & ModelNet40& ScanObjectNN  \\ 
\hline
8&92.8&79.5\\
12&93.3&80.7\\
16&\textbf{93.8}&81.4\\
20&93.6&\textbf{82.6}\\
24&93.4&81.9\\
\hline
\end{tabular}}
\end{table}

\section{Conclusion} In this paper, we propose a novel global attention-guided dual-domain feature learning network for 3D point clouds. The main contribution of our method lies in the design of the Contextual {Position-enhanced} Transformer module (CPT) and the Double K-nearest neighbor Feature Fusion (DKFF) module. {The CPT module produce a global-aware input embedding that serves as the guidance to subsequent aggregations and facilitates network learning.} The DKFF module dynamically constructs graphs in dual domains, {which appreciates both local geometric relations and long-distance semantic connections, provides highly effective feature aggregation. The proposed network can be trained in an end-to-end fashion and demonstrates superior performance on various point cloud analysis tasks, including classification, part segmentation, and large-scale semantic segmentation. Moreover, the proposed modules are ready to serve as plug-and-play blocks for downstream tasks such as completion, denoising, and compression.}


%





\ifCLASSOPTIONcaptionsoff
  \newpage
\fi




\bibliographystyle{IEEEtran}
\bibliography{mybibfile}

\end{document}